\pdfoutput=1
\documentclass{article}

\PassOptionsToPackage{numbers}{natbib}

\usepackage[final]{neurips_2023}

\usepackage[utf8]{inputenc} 
\usepackage[T1]{fontenc}    
\usepackage{hyperref}       
\usepackage{url}            
\usepackage{booktabs}       
\usepackage{amsfonts}       
\usepackage{nicefrac}       
\usepackage{microtype}      

\usepackage[ruled]{algorithm2e}

\usepackage{threeparttable}

\usepackage{graphicx}
\usepackage{subfigure}
\usepackage{booktabs} 
\usepackage{caption}
\usepackage{framed}
\usepackage{amssymb}
\usepackage{mathrsfs}
\usepackage{mathtools}
\usepackage{array}
\usepackage{amsthm}
\usepackage{verbatim} 
\usepackage{enumerate}
\usepackage{bbm}
\usepackage{commath}
\usepackage{wrapfig}
\usepackage{amsbsy}
\usepackage{float}
\usepackage{soul}
\usepackage{xcolor}
\definecolor{custom2}{HTML}{F58157}
\definecolor{custom3}{HTML}{E7434C}
\definecolor{custom4}{HTML}{99216A}
\definecolor{custom5}{HTML}{64256E}
\definecolor{custom6}{HTML}{291956}
\usepackage{amsmath}
\usepackage{tabularx}
\usepackage{listings}
\usepackage{xcolor}
\usepackage{colortbl}
\usepackage[breakable, most]{tcolorbox}
\usepackage{multirow}
\usepackage{graphicx}
\usepackage{lipsum} 
\usepackage{paralist}
\usepackage{enumitem}

\definecolor{comment}{RGB}{70, 150, 60}

\newenvironment{myitemize}{%
\begin{itemize}[leftmargin=1em, itemsep=.1em, parsep=.1em, topsep=.1em,
    partopsep=.1em]}
{\end{itemize}}

\newenvironment{myenumerate}{%
\begin{enumerate}[leftmargin=1em, itemsep=.1em, parsep=.1em, topsep=.1em,
    partopsep=.1em]}
{\end{enumerate}}

\newenvironment{structure*}{\color{blue}\begin{myenumerate}}{\end{myenumerate}}

\hypersetup{
    colorlinks,
    linkcolor={red!50!black},
    citecolor={blue!50!black},
    urlcolor={blue!80!black}
}

\lstdefinestyle{compressedstyle}{
    language=Python,
    basicstyle=\ttfamily\scriptsize,  
    breaklines=true,
    aboveskip=0pt,  
    belowskip=0pt,  
    lineskip=-2pt,  
}

\newtcblisting{compressedlisting}[2][]{
    listing only,
    arc=1mm,  
    outer arc=1mm,
    top=0mm,  
    bottom=0mm,  
    left=3mm,  
    right=3mm,  
    boxsep=2mm,  
    titlerule=0mm,  
    titlerule style={opacity=0},  
    title after break={},  
    colback=white,
    listing options={style=compressedstyle},
    title=#2,
    #1
}

\definecolor{lightorange}{RGB}{255,229,204}
\sethlcolor{lightorange}

\definecolor{lightblue}{RGB}{173,216,230}

\usepackage{sidecap}
\usepackage{wrapfig}

\newtheorem*{conjecture*}{Conjecture}
\newtheoremstyle{nonindented}{1ex}{1ex}{}{}{\bfseries}{.}{.5em}{}
\newtheoremstyle{indented}{1ex}{1ex}{\itshape\addtolength{\leftskip}{0.6cm}\addtolength{\rightskip}{0.6cm}}{}{\bfseries}{.}{.5em}{}
\theoremstyle{nonindented}
\theoremstyle{indented}
\theoremstyle{plain}

\def\max{\qopname\relax n{max}}

\def\argmax{\qopname\relax n{argmax}}

\newenvironment{lp*}{\begin{equation*}  \begin{array}{lll}}{\end{array}\end{equation*}}

\title{\framework: Learning Partially Observable World Models with LLMs for Multi-Agent Decision Making}

\author{%
  Jonathan Light$^1$\ \ Sixue Xing$^1$\ \ Yuanzhe Liu$^1$\ \  Weiqin Chen$^1$\ \ Min Cai$^2$\ \ Xiusi Chen$^3$ \\ \textbf{Guanzhi Wang$^4$}\ \   \textbf{Wei Cheng$^5$}\ \ \textbf{Yisong Yue$^4$}\ \ \textbf{Ziniu Hu$^6$}\\
  $^1$\textmd{Rensselaer Polytechnic Institute},\ \ $^2$\textmd{Shenzhen University},\\ $^3$\textmd{University of Illinois Urbana-Champaign}, $^4$\textmd{California Institute of Technology}, \\
  $^5$\textmd{NEC laboratories America}, $^6$\textmd{xAI}\\
}

\newcommand{\method}{\textsc{OmegaZero}\xspace}

\newcommand{\framework}{\textsc{PIANIST}\xspace}

\begin{document}

\maketitle

\begin{abstract}
Effective extraction of the world knowledge in LLMs for complex decision-making tasks remains a challenge. 
We propose a framework \framework for decomposing the world model into seven intuitive components conducive to zero-shot LLM generation. Given only the natural language description of the game and how input observations are formatted, our method can generate a working world model for fast and efficient MCTS simulation. 
We show that our method works well on two different games that challenge the planning and decision making skills of the agent for both language and non-language based action taking,  \emph{without any training on domain-specific training data or explicitly defined world model.}
\end{abstract}


\section{Introduction}
\vspace{-0.1in}



\begin{wrapfigure}{r}{0.4\textwidth}
  \centering
\vspace{-0.5in}  
\includegraphics[width=0.4\textwidth]{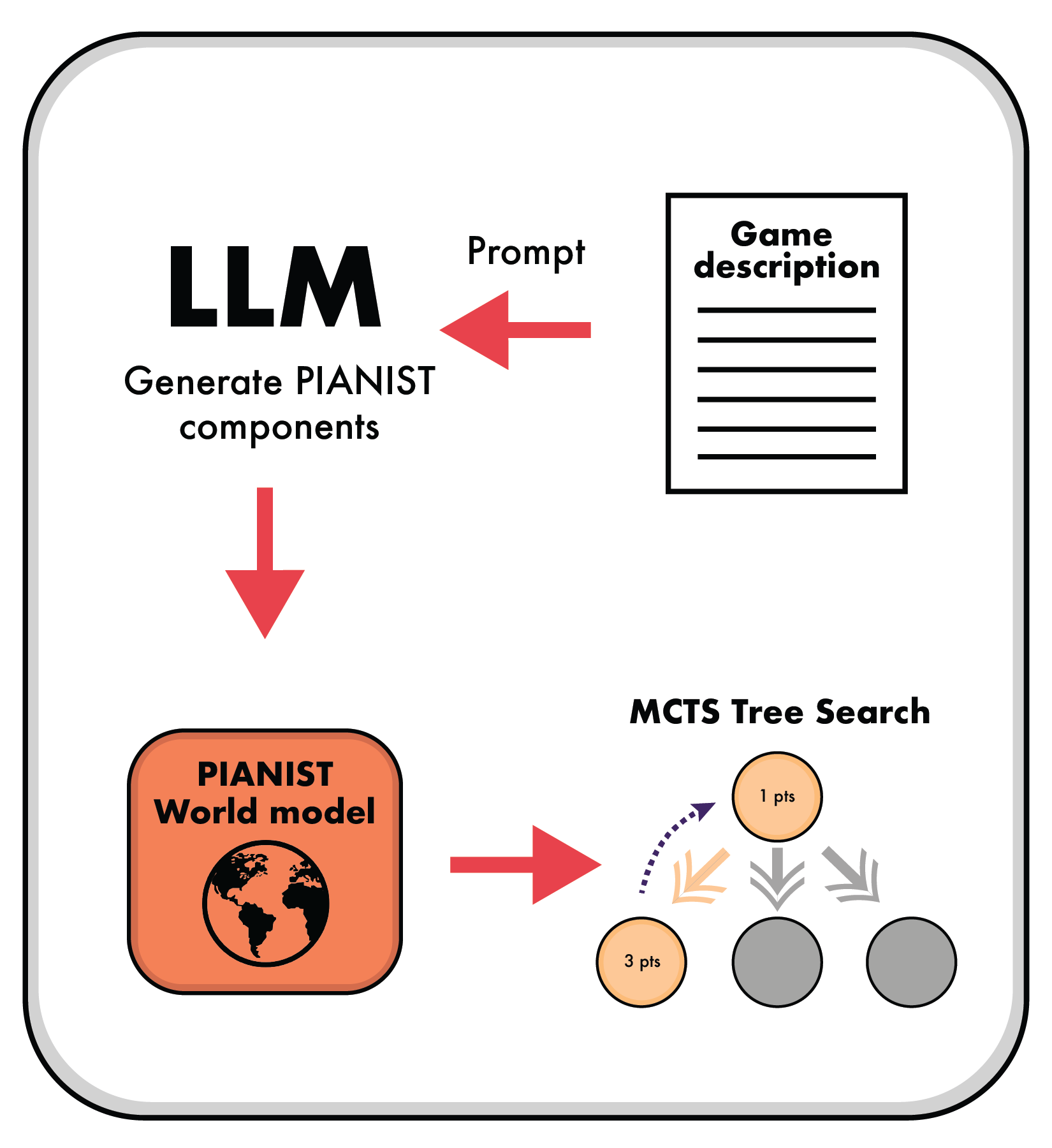}
  \caption{\textbf{Overview of \framework}. Starting with just the game description, the LLM generates a complete multi-agent, partial information world model, which can then be used for planning via search.}
  \label{fig:teasing}
\vspace{-0.2in} 
\end{wrapfigure}

Recent studies have shown how LLMs, trained on massive amounts of online data, can be used as a world model to conduct planning ~\citep{hao2023reasoning, yao2024tree}. 
However, using LLMs as world models have not been as well explored in multi-agent, partial information settings such as in language games and other board games. 
These settings present unique challenges due to (1) the complexity of all the possible action, (2) partial observablity, and (3) other, possibly adversarial or stochastic, agents. 
These complexities mean that directly using the LLM as a policy for planning is not as feasible ~\citep{zhao2024large}. More related works in App. \ref{sec:related_work}.

In this work, we introduce a framework \framework that allows us to use the LLM to more easily learn and plan with a \framework world model. 
Specifically, \framework separates the world model into seven different components that we use the LLM to generate.
This includes the forward transition function, the action function, and the information partition function, all of which we prompt the LLM to generate in the form of code, which is easily executable and verifiable. We show that our method works well on two different games -- one card based, and one discussion based -- showing strong performance from LLM-agents that use our framework.

\section{Background}
\vspace{-0.1in}
\subsection{Decision problem formulation}
\vspace{-0.1in}
We formulate decision making tasks as a partially observable Markov decision process (POMDP) with an explicit environment actor which makes it more LLM-friendly to model. 

\textbf{Problem definition.} Given a state space $\mathcal{S}$ and action space $\mathcal{A}$, a policy function $\phi$ in policy space $\Phi$ maps states to probability distributions over actions, $\phi: \mathcal{S} \rightarrow \Delta \mathcal{A}$. An environment $\mathcal{E} = \langle \mathcal{S}, \mathcal{A}, \mathcal{N}, T, R, A, \phi_{\epsilon}\rangle$ includes the state and action spaces, actors $\mathcal{N}$, a \textbf{transition function} $T: \mathcal{S} \times \mathcal{A} \rightarrow \mathcal{S}$, a \textbf{reward function} $R: \mathcal{S} \times \mathcal{A} \rightarrow \mathbb{R}^{|\mathcal{N}|}$ for actor rewards, and an \textbf{action function} $A: \mathcal{S} \rightarrow \mathcal{N}, \mathcal{P}(\mathcal{A})$ determining legal actions. The environment actor's policy $\phi_e$ handles stochastic transitions, allowing for both deterministic and stochastic, single or multi-agent settings. In partial information settings, an \textbf{information partition} function $P: \mathcal{S} \times \mathcal{N} \rightarrow \mathcal{I}$ maps hidden states to information sets. Then the policy $\phi$ maps from information sets to action distributions, $\phi: \mathcal{I} \rightarrow \Delta \mathcal{A}$.

\textbf{Goal.} Given an environment $\mathcal{E}$, the goal for each actor $i \in \mathcal{N}$ is to find a policy $\phi_i^*$ that maximizes their cumulative reward, given that other players are also playing their optimal policy $\phi_{-i}^*$: $\phi^*_i = \argmax_{\phi_i}\underset{ \tau \sim (\phi_i, \phi_{-i}^*)}{\mathbb{E}} \left[ \sum_{(s,a) \in \tau} R_i(s,a) \right]$,
where $\tau = (s_0, a_0, ...)$ is the simulated trajectory according to the strategic profile $(\phi_i, \phi_{-i})$ and the transition function $T$, with $a_t \sim \phi(a_t | s_t)$ and $s_{t+1} = T(s_t, a_t)$. $\boldsymbol{\phi}$ is commonly known as a Nash equilibrium, since no player $i$ has any incentive to individually deviate from their optimal policy $\phi_i^*$.

\subsection{Decision-making games}
\label{sec:game_description}
\vspace{-0.1in}
We evaluate the performance of \method compared to other algorithms, both rule based and deep reinforcement learning based, on two board games representing games of two different genres. 


\textbf{GOPS} (Goofspiel) is a multi-round, two-player simultaneous action game commonly studied in game theory ~\citep{lanctot2009monte, ross1971goofspiel}. Each player is dealt identical hands of cards numbered $1$ to $k$, and a shuffled prize deck, also numbered $1$ to $k$, is revealed one card at a time. Both players simultaneously play a card from their hand; the higher card wins the prize, and both cards are discarded. After $k$ rounds, players sum the values of their won prize cards, and the higher total determines the winner. Good players anticipate future moves and assess the value of each prize. \underline{Long-horizon games} challenge LLM agents, as they struggle to connect near-term actions with long-term outcomes.


\textbf{Taboo} (2-player text version) is a cooperative game where one player is the clue-master and the other is the guesser. The clue-master is given a target word and a list of taboo words they cannot use in their clues. Each round, the clue-master makes a statement, and the guesser responds with one guess. The game ends when the guesser correctly identifies the word, makes five guesses, or the clue-master accidentally uses a taboo word. If a taboo word is used, the team scores 0; otherwise, the score is five minus the number of guesses. A good clue-master \emph{anticipates the guesser's thought process} to help narrow down the options. The novelty of each word adds to the challenge.

\section{Methodology}
\label{sec:policyagent}
\vspace{-0.1in}

\subsection{\framework: Extracting LLM World Knowledge}
We present a new framework for extracting world knowledge from LLMs by dividing the world model into seven intuitive components that the LLM can understand. With this extracted model, we can apply model-based reinforcement learning techniques like MCTS or TD-learning. Most components are generated by prompting the LLM with the game description and a predefined Python parent template class. See App. \ref{sec:pianist_examples} for examples of LLM generated models. 
\begin{itemize}[nosep,leftmargin=*] 
\item \textbf{$\mathcal{I}$: Information sets.} The agent observes information sets, and we provide code for representing them along with a natural language game description as an interface between the real world and the agent. This and the game description are the only game-specific information given. 
\item \textbf{$\mathcal{S}$: Hidden states.} The agent records any relevant hidden information here. 
\item \textbf{$\mathcal{N}$: Actors.} 
Used to specify what the actor names are for the action function and reward function.
\item \textbf{$A$: Action function.} For large action spaces, the function returns the top $k$ most likely actions. For language actions, an LLM generates the top $k$ text options. See section \ref{sec:language_actions} for details. 
\item \textbf{$T, R$: Transition-reward function.} Combining state prediction and reward assignment for each player minimizes LLM errors. Deterministic transitions further reduce generation errors. 
\item \textbf{$P$: Information partition function.} 
\item \textbf{$I: \mathcal{I} \rightarrow \mathcal{S}$: Information realization function.} This maps information sets to their most likely hidden states, enabling the agent to simulate transitions between hidden states. \end{itemize}
Together, we have the \textbf{P}artition function, \textbf{I}nformation set space, \textbf{A}ction space function, \textbf{N} players, \textbf{I}nformation realization function, \textbf{S}tate space, and \textbf{T}ransition-reward function, or \framework for short. If the LLM generates incorrect code, we use a reflexion approach to correct it and regenerate ~\citep{madaan2024self, shinn2024reflexion}. We choose to learn hidden states and transitions because it's more intuitive for the LLM to understand actions and transitions at hidden states rather than at information sets.

\subsection{Integrating \framework with Search}
\begin{figure}[h]
\vspace{-0.1in}
    \centering
    \includegraphics[width = 1.0\textwidth]{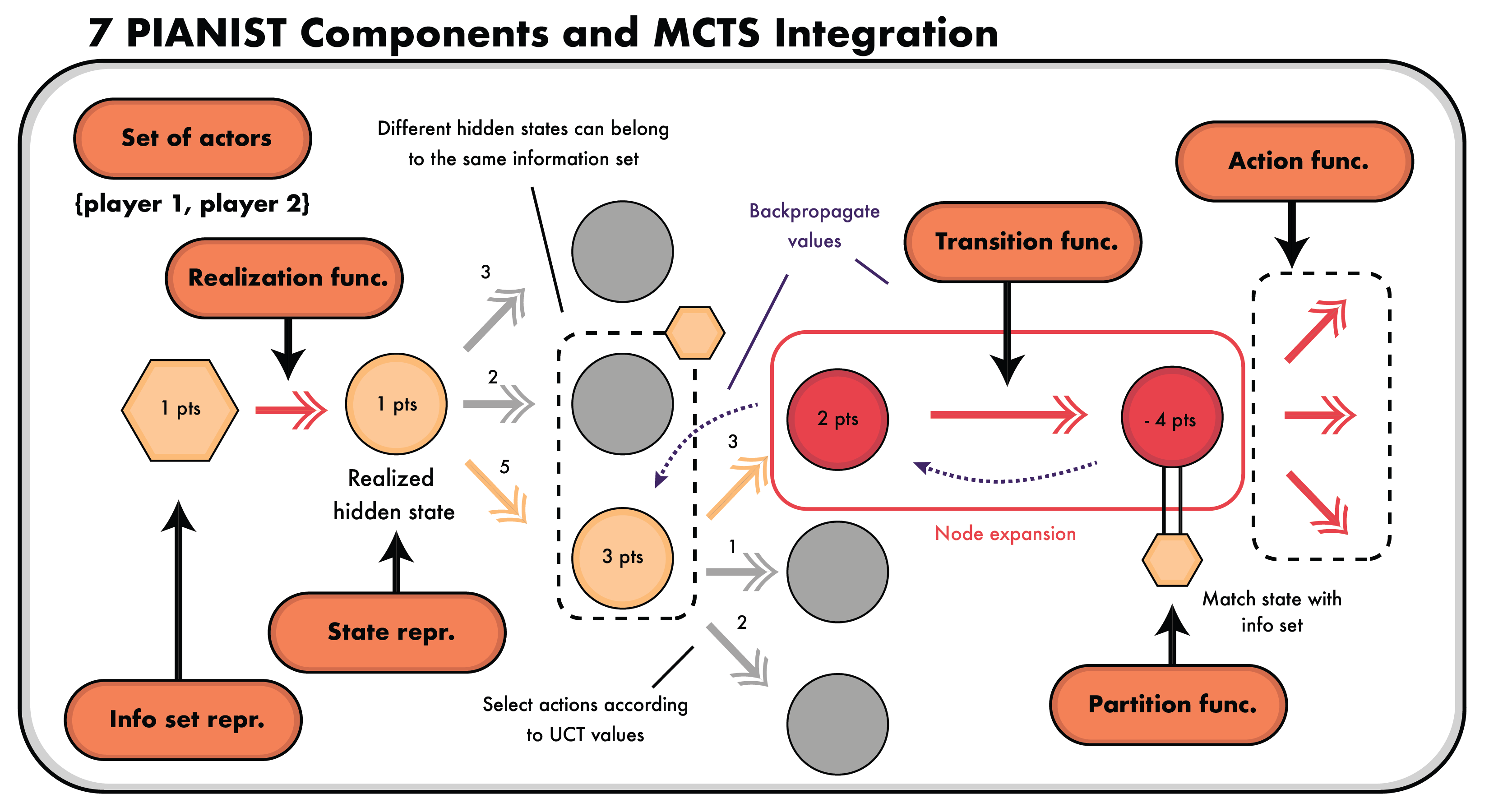}
    \vspace{-0.2in}
    \caption{
    \textbf{Integrating \framework components with MCTS}. The realization function samples a hidden state for simulation, while the transition, action, and partition functions are used to expand new states. States are selected based on UCT values, aggregated across information sets for partial information. Though the diagram shows values for a single player, in practice, values for all players are inferred and updated simultaneously. See App. \ref{sec:mcts_details} for details and Fig. \ref{fig:generation_graph} for generation order. 
    }
    \label{fig:mcts_int}
    \vspace{-0.1in}
\end{figure}

\vspace{-0.1in}
Our pseudocode for \framework-guided MCTS is in Alg. \ref{alg:mcts}, with a diagram in Fig. \ref{fig:mcts_int}, and further details in App. \ref{sec:mcts_details}. During MCTS, we sample a realization of the observed information set using the information realization function. A trajectory is then simulated by selecting the action with the highest UCT value (eq. \ref{eq:uct}) for the acting player. Since actors cannot distinguish hidden states within the same information set, UCT values are averaged across all states in that set, weighted by visit counts, preventing the use of hidden information. The simulation continues until a state $\boldsymbol{s}$ with unexplored actions is reached, where a random unexplored action $\boldsymbol{a}$ is chosen. The transition function provides the next state $\boldsymbol{s}'$ and reward $\boldsymbol{r}$, which are recorded. The action function, partition function, and value heuristic are used to record the actions, information set, and value estimate for $\boldsymbol{s}'$. A random rollout or LLM-generated heuristic computes the value estimate. Backpropagation is then performed from $\boldsymbol{s}'$ up the tree using the backpropagation equation (eq. \ref{eq:backprop}).


\subsection{Handling language actions}
\label{sec:language_actions}
\vspace{-0.1in}

\underline{Language-based games} are particularly challenging for traditional RL methods due to their need for \emph{language abilities and extensive semantic knowledge}. In these games, the action space for language-based actions, such as discussion, is practically infinite, consisting of all possible word and token combinations. Current search methods are only effective in finite action spaces. Additionally, RL methods alone cannot inherently understand language or be trained to do so through self-play. We address this by utilizing LLMs to propose likely high-level dialogue actions for players. This allows us to (1) prune improbable actions and (2) focus on a few high-level strategy categories, reducing the search space. The LLM only suggests possible actions, while the search algorithm assigns probabilities, mitigating the bias issue commonly found in LLM decision making ~\citep{xu2023language}.

\section{Experiments}
\vspace{-0.1in}


We evaluated our model against three different opponents. For the \textbf{ground-truth models}, we used ground truth models with MCTS search, combined with a random-rollout value heuristic, and played them against our LLM-generated agent, which also uses an LLM-generated value heuristic. Ground-truth include the true $\mathcal{S}, \mathcal{A}, \mathcal{N}, T, R, A$ models used during actual gameplay. For \textbf{LLM as policy}, we directly queried the LLM for actions in ReAct style ~\citep{yao2022react}, which includes a thought phase before action. For \textbf{human opponents}, we recruited 10 individuals to play 30 games of 6-card GOPS and 30 games of Taboo. In Taboo (a cooperative game), we paired each agent with a human-crafted model as the teammate (guesser), as the clue-giver role is more difficult. In GOPS, the two agents played directly against each other. We report both win rate and score for both games. In GOPS, win rate refers to whether a player had a higher score than their opponent, while score represents the point difference based on how many score cards were won. In Taboo, win rate measures whether the team guessed the word on the first try, and score is based on how quickly the team guessed the correct word. Note the possibility for tieing in GOPS. 


As shown in Table \ref{tab:truth_exp}, \framework performs similarly to ground-truth models, indicating that the LLM can generate an accurate world model using our framework. Additionally, Table \ref{tab:policy_exp} shows that our world model helps the agent plan more effectively than directly querying the LLM for actions. However, our agent struggles to consistently beat humans at GOPS and Taboo, highlighting the need for further research on improving LLM agents for complex decision-making environments on both action and language games (Table \ref{tab:human_exp}). Overall, despite using LLMs to generate its world model zero-shot, our agent demonstrates strong performance, showcasing the effectiveness of \framework in extracting world knowledge for tree search. This suggests that future work could explore more nuanced adaptations of the framework to balance decision-making performance across different games, potentially enabling more robust generalization in varied multi-agent environments.

\begin{table}[ht]
\caption{\textbf{\framework vs Ground truth models,} comparing performance when we replace ground truth models with LLM generated models.}
\label{tab:truth_exp}
\centering
\begin{tabular}{@{}cccccccc@{}}
\toprule
\multirow{2}{*}{\textbf{Game}} & \multirow{2}{*}{\textbf{Setting}} & \multirow{2}{*}{\textbf{\# games}} & \multicolumn{2}{c}{\textbf{\framework}} & \multicolumn{2}{c}{\textbf{Ground-truth}} \\ 
\cmidrule(lr){4-5} \cmidrule(lr){6-7}
& & & Winrate & Score & Winrate & Score \\
\midrule
\multirow{2}{*}{GOPS} & 6-card & 300 & 52.3$\pm$6.2\% & -0.21 & 48.7$\pm$1.9\% & 0.21 \\
                 & 12-card   & 300 & 47.3$\pm$2.9\% & -0.07 & 47.0$\pm$5.7\% & 0.07 \\ \midrule
\multirow{1}{*}{Taboo}  & as clue giver & 15 & 60.0$\pm$8.7\% & 3.53 & 53.3$\pm$12.9\% & 3.06\\

\bottomrule
\end{tabular}
\end{table}



\begin{table}[ht]
\caption{\textbf{\framework vs LLM as Policy}, where the LLM directly chooses which actions to take ~\citep{yao2022react}.}
\label{tab:policy_exp}
\centering
\begin{tabular}{@{}cccccccc@{}}
\toprule
\multirow{2}{*}{\textbf{Game}} & \multirow{2}{*}{\textbf{Setting}} & \multirow{2}{*}{\textbf{\# games}} & \multicolumn{2}{c}{\textbf{\framework}} & \multicolumn{2}{c}{\textbf{LLM-policy}} \\ 
\cmidrule(lr){4-5} \cmidrule(lr){6-7}
& & & Winrate & Score & Winrate & Score \\
\midrule
\multirow{2}{*}{GOPS} & 6-card & 30 & 66.6$\pm$9.4\% & 0.3 & 33.3$\pm$9.4\% & -0.3\\
                 & 12-card   & 30 & 60.0$\pm$8.2\% & 0.2 & 33.3$\pm$9.4\% & -0.2 \\ \midrule
\multirow{1}{*}{Taboo}  & as clue giver & 15 & 60.0$\pm$8.7\% & 3.53 & 53.3$\pm$12.9\% & 3.2 & \\
\bottomrule
\end{tabular}
\end{table}

\begin{table}[ht]
\caption{\textbf{\framework vs the Humans}. Humans were given the same prompt as the LLM before play.}
\label{tab:human_exp}
\centering
\begin{tabular}{@{}cccccccc@{}}
\toprule
\multirow{2}{*}{\textbf{Game}} & \multirow{2}{*}{\textbf{Setting}} & \multirow{2}{*}{\textbf{\# games}} & \multicolumn{2}{c}{\textbf{\framework}} & \multicolumn{2}{c}{\textbf{the humans}} \\ 
\cmidrule(lr){4-5} \cmidrule(lr){6-7}
& & & Winrate & Score & Winrate & Score \\
\midrule
\multirow{1}{*}{GOPS} & 6-card & 40 & 54.9$\pm$8.3\% & 2.37 & 40.1$\pm$7.1\% & -2.41 \\
\midrule
\multirow{1}{*}{Taboo}  & as clue giver & 15 & 60.0$\pm$8.7\% & 3.53 & 86.7$\pm$8.7\% & 4.33\\
\bottomrule
\end{tabular}
\end{table}

\newpage
\clearpage

\bibliography{ref}

\clearpage
\appendix

\section{MCTS Details}
\label{sec:mcts_details}

The Monte Carlo Tree Search (MCTS) process, illustrated in Figure \ref{fig:mcts_int}, simulates possible future game states by expanding nodes in a search tree. Each node corresponds to a state, and edges correspond to actions. MCTS operates by iteratively simulating trajectories (denoted as $\tau$) from the current realized hidden state until an unexpanded state is reached. The steps involved in MCTS include four key phases: selection, expansion, simulation, and backpropagation.

\subsection{Upper Confidence Bound for Trees (UCT) and Action Selection}
At each node, the next action $\boldsymbol{a}^*$ is selected according to the Upper Confidence Bound for Trees (UCT) formula, given by Equation \eqref{eq:uct}:

\begin{equation}
    \label{eq:uct}
    \boldsymbol{a}^* = \arg\max_{\boldsymbol{a}} \left( \underset{\boldsymbol{s} \in I_{\boldsymbol{s}}}{\mathbb{E}}\left[ r_i + \gamma V_i(\boldsymbol{s}') + C \sqrt{\frac{\log n(\boldsymbol{s})}{n(\boldsymbol{s}')}} \mid \boldsymbol{s}', \boldsymbol{r} = T(\boldsymbol{s}, \boldsymbol{a})\right] \right)
\end{equation}

In this equation:

\begin{itemize}
    \item $\boldsymbol{a}^*$ is the action that maximizes the expression.
    \item $\boldsymbol{s}$ represents the current state, and $\boldsymbol{s}'$ is the next state reached after taking action $\boldsymbol{a}$.
    \item $r_i$ is the immediate reward for agent $i$ when transitioning from state $\boldsymbol{s}$ to $\boldsymbol{s}'$.
    \item $V_i(\boldsymbol{s}')$ is the value function that estimates the future reward from the next state $\boldsymbol{s}'$ for agent $i$.
    \item $\gamma$ is the discount factor, which balances immediate and future rewards.
    \item $n(\boldsymbol{s})$ is the number of visits to the current state $\boldsymbol{s}$, while $n(\boldsymbol{s}')$ is the number of visits to the next state $\boldsymbol{s}'$.
    \item $C$ is a constant controlling exploration versus exploitation, and $\log n(\boldsymbol{s}) / n(\boldsymbol{s}')$ encourages exploring less-visited actions.
    \item $T(\boldsymbol{s}, \boldsymbol{a})$ represents the transition dynamics, mapping the current state and action to the next state and reward.
\end{itemize}

\subsection{Probability Distribution Over the Information Set}
Since our MCTS handles partial observability, a probability distribution is defined over the information set $I_{\boldsymbol{s}}$, which contains all possible hidden states $\boldsymbol{s}$ that are consistent with the observed information. This distribution is given by:

\begin{equation}
    \mathbb{P}(\boldsymbol{s}) = \frac{n(\boldsymbol{s})}{\sum_{\boldsymbol{s}' \in I_{\boldsymbol{s}}} n(\boldsymbol{s}')}
\end{equation}

Where:

\begin{itemize}
    \item $\mathbb{P}(\boldsymbol{s})$ is the probability of being in state $\boldsymbol{s}$ within the information set $I_{\boldsymbol{s}}$.
    \item $n(\boldsymbol{s})$ is the visit count of state $\boldsymbol{s}$, and the denominator normalizes over all possible states $\boldsymbol{s}' \in I_{\boldsymbol{s}}$.
\end{itemize}

\subsection{Backpropagation}
After simulating a trajectory $\tau$, MCTS backpropagates the results of the simulation to update the value estimates and visit counts for all state-action pairs $(\boldsymbol{s}, \boldsymbol{s}')$ along the trajectory. This update is performed using Equation \eqref{eq:backprop}:

\begin{equation}
    \label{eq:backprop}
    \forall i \in \mathcal{N} \quad V_i(\boldsymbol{s}) \leftarrow V_i(\boldsymbol{s}) + \frac{1}{n(\boldsymbol{s})} \left( r_i + \gamma V_i(\boldsymbol{s}') - V_i(\boldsymbol{s}) \right), \quad n(\boldsymbol{s}) \leftarrow n(\boldsymbol{s}) + 1
\end{equation}

\newpage
\begin{algorithm}[H]
\SetAlgoLined
\KwIn{Initial information set $\boldsymbol{h}_0$, number of iterations $M$}
\KwOut{Best action $\boldsymbol{a}^*$}

\SetKwFunction{FSelect}{Select}
\SetKwFunction{FExpand}{Expand}
\SetKwFunction{FValue}{EstimateValues}
\SetKwFunction{FBackpropagate}{Backpropagate}
\SetKwFunction{FBestChild}{BestChild}
\SetKwFunction{FInformationSet}{InformationSet}
\SetKwFunction{FRealize}{Realize}

\SetKwProg{Fn}{Function}{:}{}
\Fn{\FSelect{$node$}}{
    \While{node is non-terminal}{
        \If{all node.actions have been tried}{
            $node \gets$ \FBestChild{node}\;
        }\Else{
            $\boldsymbol{a} = $ random untried action\;
            \Return $node, \boldsymbol{a}$\;
        }
    }
    \Return $node$ (terminal state)\;
}

\Fn{\FExpand{$parent, \boldsymbol{a}$}}{
    $\boldsymbol{s} \gets T(parent.s, \boldsymbol{a})$\;
    Create new node $child$ with parent $parent$ and $child.s\gets\boldsymbol{s}$\;
    $child.actions, child.actor \gets A(\boldsymbol{s})$\;
    $child.h \gets P(\boldsymbol{s})$\;
    $child.values \gets$ \FValue($child$)\;

    \Return $child$\;
}

\Fn{\FValue{$node$}}{
    Use random rollout or some other value heuristic to estimate the value of state $node.s$ for each player\;
    \Return estimated values\;
}

\Fn{\FBackpropagate{$node$}}{
    $next\_values = node.values$\;
    \While{$node$ is not null}{
        $node \gets$ parent of $node$\;
        Update $node.values$ with the $next\_values$\;
    }
}

\Fn{\FBestChild{$root$}}{
    \Return child of $root$ with highest average reward\;
}

\Fn{\FInformationSet{$node$, $\boldsymbol{h}$}}{
    Generate the updated information set $\boldsymbol{h'}$ for $node$ based on observed actions and outcomes within $\boldsymbol{h}$\;
    \Return $\boldsymbol{h'}$\;
}
\Fn{\FRealize{$\boldsymbol{h}$}}{
    $nodes \gets $ all nodes in graph with $node.h = \boldsymbol{h}$\;
    \If{$nodes$ is empty}{
        Create a new node $node$ with $node.h = \boldsymbol{h}$ and $node.s=I(\boldsymbol{h})$ and add to graph
    }
    \Else{
        $node \gets$ RandomChoice($nodes$)\;
    }
    \Return $node$\;
}
\SetKwBlock{Begin}{begin}{end}

\Begin{

    \For{$i = 1$ \KwTo $N$}{
        $root \gets$ \FRealize{$\boldsymbol{h}_0$}\;
        $node, \boldsymbol{a} \gets$ \FSelect{$root$}\;
        $child \gets$ \FExpand{$node, \boldsymbol{a}$}\;
        \FBackpropagate{$child$}\;
    }
    $\boldsymbol{a}^* \gets$ \FBestChild{$root$}\;
}
\caption{Monte Carlo Tree Search for Partial Information with Information Sets}
\label{alg:mcts}
\end{algorithm}

\newpage
Here:

\begin{itemize}
    \item $V_i(\boldsymbol{s})$ is updated for agent $i$ by adding a fraction of the difference between the expected future reward (given by $r_i + \gamma V_i(\boldsymbol{s}')$) and the current value estimate $V_i(\boldsymbol{s})$.
    \item $r_i$ is the reward obtained from transitioning between $\boldsymbol{s}$ and $\boldsymbol{s}'$.
    \item $\gamma$ is the discount factor, which balances immediate and future rewards.
    \item $n(\boldsymbol{s})$ is incremented to reflect that the state $\boldsymbol{s}$ has been visited once more.
\end{itemize}

This backpropagation process ensures that the value estimates $V_i(\boldsymbol{s})$ are refined based on simulated outcomes, allowing the MCTS process to converge on more accurate policies over time.

By iteratively simulating trajectories, selecting actions, expanding nodes, and backpropagating rewards, MCTS effectively balances exploration and exploitation, making it a powerful search algorithm for solving decision-making problems in partially observable environments.


\section{\framework Generation details}
\begin{figure}[h!]
    \centering
    \includegraphics[width = 1.0\textwidth]{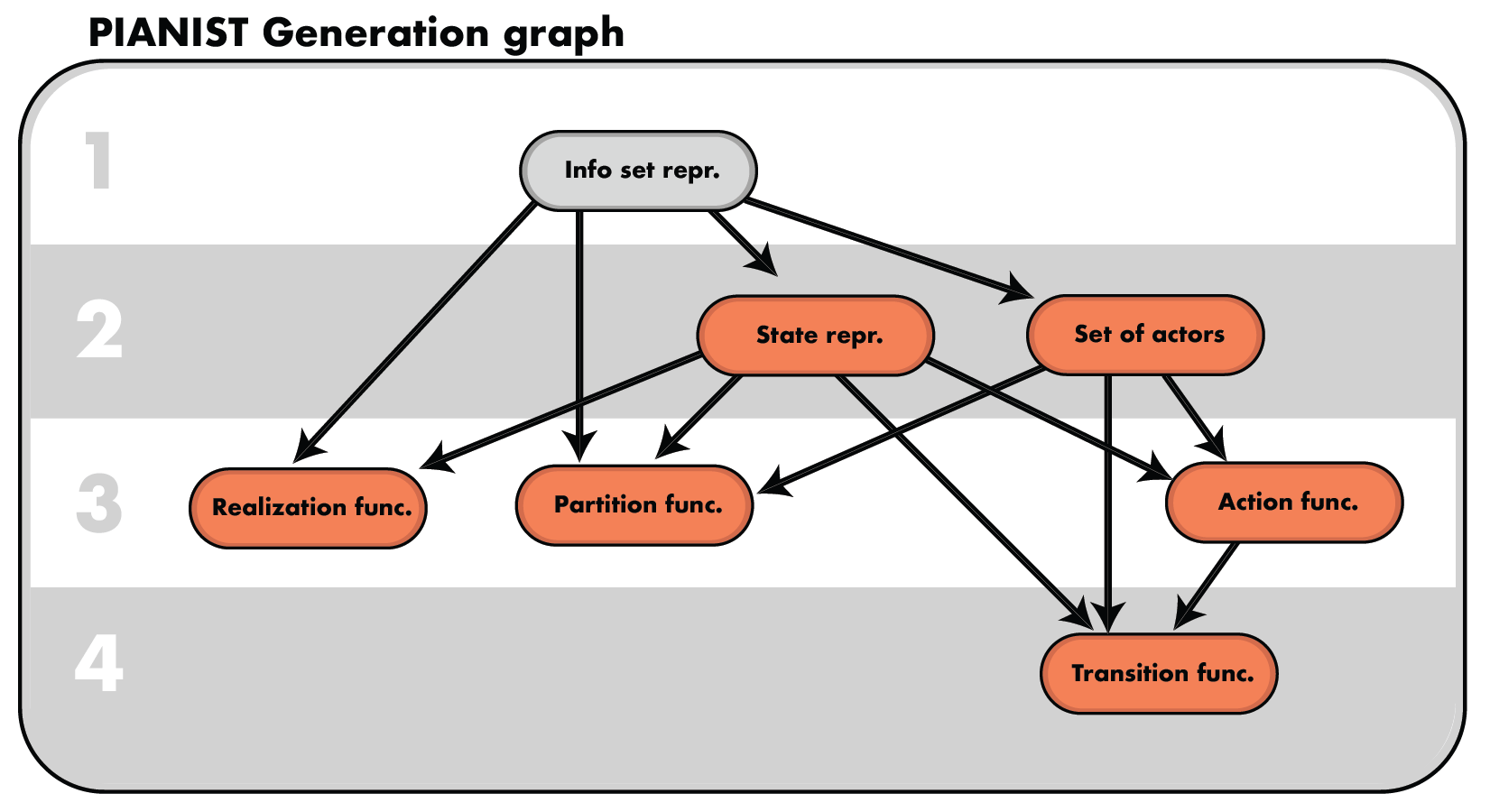}
    \caption{
    \textbf{Directed generation graph for \framework}. We display the sequential generation order for the various components of \framework, with dependencies shown by directed arrows. Generating and testing objects in this order minimizes the probability of execution failure. The initial information set representation is given by the environment to allow an unified interface with the environment. Modularization also means we can test each component individually. 
    }
    \label{fig:generation_graph}
\end{figure}

\newpage

\section{Example \framework models generated by \texttt{gpt-4o}}
\label{sec:pianist_examples}

\begin{compressedlisting}[breakable, enhanced]{Example LLM Generated Forward Dynamics Model (GOPS)}
class CustomForwardTransitor(ForwardTransitor):
    '''
    Custom forward transitor for the Game of Pure Strategy (GOPS).
    Implements the game logic for transitioning between states.
    '''

    def _transition(self, state: HiddenState, action: int, actor: int) -> Tuple[HiddenState, Dict[int, float]]:
        '''
        Transits to the next state given the current state, actor, and action taken by the actor.

        Args:
            state: current HiddenState
            action: card played by the actor
            actor: actor that is taking the action (0 for player 0, 1 for player 1)

        Returns:
            next_state: updated HiddenState
            rewards: reward of the transition for each player
        '''
        # Copy current state variables to modify them
        player_0_hand = state.player_0_hand
        player_1_hand = state.player_1_hand
        prize_deck = state.prize_deck
        player_0_played_cards = state.player_0_played_cards
        player_1_played_cards = state.player_1_played_cards
        played_prize_cards = state.played_prize_cards
        player_0_cumulative_score = state.player_0_cumulative_score
        player_1_cumulative_score = state.player_1_cumulative_score
        contested_points = state.contested_points

        # Record the action taken by the actor
        if actor == 0:
            player_0_played_cards += (action,)
            player_0_hand = player_0_hand - frozenset([action])
        elif actor == 1:
            player_1_played_cards += (action,)
            player_1_hand = player_1_hand - frozenset([action])

        # If both players have played cards, determine the winner
        if player_0_played_cards and player_1_played_cards:
            player_0_card = player_0_played_cards[-1]
            player_1_card = player_1_played_cards[-1]
            
            if player_0_card > player_1_card:
                # Player 0 wins the round
                if contested_points > 0:
                    player_0_cumulative_score += contested_points + played_prize_cards[-1]
                    contested_points = 0
                else:
                    # Award the prize card if it exists
                    if prize_deck:
                        prize_card = prize_deck.pop()
                        player_0_cumulative_score += prize_card
                        played_prize_cards += (prize_card,)
                    else:
                        contested_points = 0  # No prize card to award
            elif player_1_card > player_0_card:
                # Player 1 wins the round
                if contested_points > 0:
                    player_1_cumulative_score += contested_points + played_prize_cards[-1]
                    contested_points = 0
                else:
                    # Award the prize card if it exists
                    if prize_deck:
                        prize_card = prize_deck.pop()
                        player_1_cumulative_score += prize_card
                        played_prize_cards += (prize_card,)
                    else:
                        contested_points = 0  # No prize card to award
            else:
                # Cards are equal, points are contested
                contested_points += 1

        # Prepare rewards dictionary
        rewards = {0: player_0_cumulative_score, 1: player_1_cumulative_score}

        # Initialize new state at the point of return
        new_state = HiddenState(
            player_0_hand=player_0_hand,
            player_1_hand=player_1_hand,
            prize_deck=prize_deck,
            player_0_played_cards=player_0_played_cards,
            player_1_played_cards=player_1_played_cards,
            played_prize_cards=played_prize_cards,
            player_0_cumulative_score=player_0_cumulative_score,
            player_1_cumulative_score=player_1_cumulative_score,
            contested_points=contested_points
        )

        return new_state, rewards
\end{compressedlisting}

\begin{compressedlisting}{Example LLM Generated Value Heuristic Function (GOPS)}
class CustomValueHeuristic(ValueHeuristic):
    '''
    Custom value heuristic for evaluating the state of the GOPS game.
    This heuristic estimates the expected rewards based on the cards in hand
    and prize cards available.

    Attributes:
        player_0_weight: Weight given to Player 0's score.
        player_1_weight: Weight given to Player 1's score.
    '''

    def _evaluate(self, state: HiddenState) -> tuple[dict[int, float], dict]:
        '''
        Evaluates the state

        Args:
            state: current hidden state of the game

        Returns:
            values: estimated values of the state for each player
            notes: additional notes about the evaluation
        '''
        player_0_value = self._estimate_value(state.player_0_hand, state.prize_deck)
        player_1_value = self._estimate_value(state.player_1_hand, state.prize_deck)

        values = {
            0: player_0_value + state.player_0_cumulative_score,
            1: player_1_value + state.player_1_cumulative_score
        }

        notes = {
            'player_0_hand': state.player_0_hand,
            'player_1_hand': state.player_1_hand,
            'prize_deck': state.prize_deck,
            'player_0_value': player_0_value,
            'player_1_value': player_1_value
        }

        return values, notes

    def _estimate_value(self, hand: FrozenSet[int], prize_deck: FrozenSet[int]) -> float:
        '''
        Estimates the potential value of a player's hand based on available prize cards.

        Args:
            hand: The player's hand of cards.
            prize_deck: The current prize deck cards.

        Returns:
            estimated value for the player's hand.
        '''
        if not prize_deck:
            return 0.0  # No prize cards left to claim

        # Count the number of prize cards that can be won
        prize_values = [prize for prize in prize_deck if prize in hand]

        # If the player has cards that can win the prize cards, sum their values
        value = sum(prize_values)
        
        # Optionally, apply a strategic factor or heuristic for more complex evaluations
        return value
\end{compressedlisting}

\begin{compressedlisting}{Example LLM Generated Actor-Action Selection Model (GOPS)}
class CustomActorActionEnumerator(ActorActionEnumerator):
    '''
    Custom enumerator for the Game of Pure Strategy (GOPS) that determines which player can act and what actions they can take.
    '''

    def _enumerate(self, state: Hashable) -> tuple[Optional[int], frozenset]:
        '''
        Enumerates the acting player and their available actions based on the current game state.
        The first player in the player order is always returned.

        Args:
            state: current state (expected to be a HiddenState)

        Returns:
            actor: the acting actor (either 0 or 1 for players).
            actions: set of actions that the actor may take (cards they have left in their hand).
        '''
        if not isinstance(state, HiddenState):
            raise ValueError("State must be an instance of HiddenState.")

        # Determine the acting player (always the first player in the order for this game)
        actor = self.player_order[0]

        # Get the actions available to the acting player
        if actor == 0:
            actions = state.player_0_hand
        else:
            actions = state.player_1_hand

        return actor, frozenset(actions)

\end{compressedlisting}

\begin{compressedlisting}{Example LLM Generated Information Realization Function (GOPS)}
class CustomInformationPrior(InformationPrior):
    '''
    Custom implementation of InformationPrior for the Game of Pure Strategy (GOPS).

    This class maps an information set, which may represent a state of the game where players have played certain cards, to a corresponding hidden state.
    '''
    
    def _get_prior_state(self, information_set: Hashable) -> HiddenState:
        '''
        Returns a hidden state corresponding to the given information set. This implementation generates a hidden state based on the current observed state.
        
        It assumes that the information_set contains relevant details such as played cards and hands.
        '''
        # Placeholder for actual extraction of information from the information set.
        # The actual implementation would depend on the structure of information_set.
        
        player_0_hand = information_set.player_0_hand
        player_1_hand = information_set.player_1_hand
        prize_deck = information_set.prize_deck
        player_0_played_cards = information_set.player_0_played_cards
        player_1_played_cards = information_set.player_1_played_cards
        played_prize_cards = information_set.played_prize_cards
        player_0_cumulative_score = information_set.player_0_cumulative_score
        player_1_cumulative_score = information_set.player_1_cumulative_score
        contested_points = information_set.contested_points
        
        # Create and return the hidden state
        return HiddenState(
            player_0_hand=player_0_hand,
            player_1_hand=player_1_hand,
            prize_deck=prize_deck,
            player_0_played_cards=player_0_played_cards,
            player_1_played_cards=player_1_played_cards,
            played_prize_cards=played_prize_cards,
            player_0_cumulative_score=player_0_cumulative_score,
            player_1_cumulative_score=player_1_cumulative_score,
            contested_points=contested_points,
        )
        
\end{compressedlisting}
\newpage

\section{Related Work}
\label{sec:related_work}


\textbf{LLMs for text agents}. Large language models (LLMs) have demonstrated significant emergent capabilities, such as zero-shot prompting and complex reasoning~\citep{bommasani2021opportunities,brown2020gpt3,raffel2020t5,wei2022emergent,chowdhery2022palm,chung2022flan}. They also possess extensive world knowledge~\citep{yu2023kola}, which has spurred increasing efforts to use LLMs for decision-making in text agents~\citep{wang2024survey}. One notable paradigm is ReAct~\citep{yao2023react}, which employs an observation-reasoning-acting loop for agent planning with LLMs. Building on ReAct, Reflexion~\citep{shinn2024reflexion} incorporates self-reflection to enhance reasoning capabilities. Other works in this domain have utilized feedback~\citep{wang2023voyager,huang2022inner}, memory~\citep{park2023generative}, and tool use~\citep{schick2024toolformer,cai2023large} to further enhance agent performance. Our proposed method, \method, integrates these components to design an agent capable of systematic analysis and strategic decision-making. Typical prompting techniques for text agents include Chain-of-Thought~\citep{wei2022chain}, Tree-of-Thought~\citep{yao2024tree}, and Graph-of-Thought~\citep{besta2024graph}. 


\textbf{LLMs and planning}. Recent works have proposed planing using the LLM as a world model ~\citep{hao2023reasoning, hu2023language}. These works have mostly centered around using the LLM as a forward transition function (dynamics model) by querying the LLM for the next state ~\citep{yao2024tree}, or using a planning language to describe plans ~\citep{guan2023leveraging}. Other works have explored using LLMs to guide the MCTS search process by using the LLM as a policy ~\citep{zhao2024large, feng2023alphazero}. We build upon these works by investigating what kind of world model is more conducive to extracting world knowledge from the LLM and combining it with MCTS. 


\textbf{Skill learning with LLMs}. Recent works have explored the possibly of LLMs learning skills through learning a textual short and long term memory ~\citep{shinn2024reflexion, majumder2023clin}, or  textual insights extracted from the memories ~\citep{zhao2024expel}. 
Due to the length of trajectories in our game setting and the numerical nature of the data, it is difficult to learn textual memories, so we learn high level strategies instead. 
We also explore how to acquire simulational self-play feedback in multiagent settings. 
Using LLMs to learn a functional reward model has also been applied to great success on single-agent robotic tasks ~\citep{ma2023eureka, yu2023language}. 
We build upon their work by introducing a new improvement method that can help learn a better reward model, and exploring how function learing can be applied to multiagent settings with simulated feedback.
\textbf{AI in strategy games.}
AI has been applied to great success in board games. AlphaGo and MuZero demonstrated the power of combining MCTS, deep learning, and feedback generation using self-play in games such as Go, Chess, and Shogi~\citep{silver2017mastering,schrittwieser2020mastering}. 
Language models can also be trained on human in-game discussion data and integrated with another separately trained action planner to play board games with dialogue ~\citep{meta2022human}. 
We build upon the AI for games literature by showing that LLMs can accomplish both \emph{ (1) the training of a value heuristic like that in AlphaGo through self-play more efficiently than RL and (2) dialogue generation in discussion games with no human examples}. 
These adversarial environments are not just limited to board games. For example, there has been recent interest on creating LLM-agents that can negotiate~\citep{abdelnabi2023llm,fu2023improving}, which our method can also be applied to. 
Traditionally the solution to searching over a large action space has been to bucket similar actions together, such as possible raises in poker ~\citep{brown2019superhuman}. 
We leverage the inherent distribution in the LLM to suggest the top most probable, yet distinct, actions instead. 

\newpage

\section{Prompts used to generate components}

\begin{compressedlisting}{P: Information Function Generation Prompt (Taboo)}

SYSTEM PROMPT:

"""
You are a programmer developing an accurate game engine. Your task is to implement parts of the game simulator in Python. The simulator models simultaneous actions as sequential ones with partial observation. When players $1, ..., k$ take simultaneous actions, they do so sequentially without seeing the actions of previous players. These actions are first recorded in a `HiddenState` object before being revealed. Do not repetitively generate Hidden state. If new state is generated, construct at return statement. `ObservedState` and `HiddenState` should be under `@dataclass(frozen=True)`.  Use tuple instead of list to make sure that vectors are frozen.
"""

HUMAN PROMPT:

from typing import Optional, Tuple
    clue_word: str                               # The actual clue word (hidden from guesser)
    taboo_words: Tuple[str, ...]                 # Taboo words (hidden from guesser)
    taboo_word_used: bool                         # Whether a taboo word was used
    guesses: Tuple[str, ...]                      # Guesses made (hidden from clue-master)
    clue_master_statements: Tuple[str, ...]      # Statements made (hidden from guesser)

"""
An observed state (information set) in the game is defined as follows:
"""
@dataclass(frozen=True)
class ObservedState:
    clue_word: Optional[str]                     # The word the guesser needs to guess
    taboo_words: Optional[tuple[str, ...]] # List of taboo words the clue-master cannot use
    guesses: tuple[str, ...] = tuple()  # List of words guessed by the guesser
    clue_master_statements: tuple[str, ...] = tuple()  # Statements made by the clue-master
    taboo_word_used: bool = False      # Flag to indicate if a taboo word was used
    game_over: bool = False            # Flag to indicate if the game is over
    score: int = 5                     # Initial score, will decrease based on guesses
    actor: str = "clue_master"          # Indicates whose turn it is: "clue_master" or "guesser"

"""
Write an information function `CustomInformationFunction` for this game that inherits from the `InformationFunction` class. Include all docstings from the parent class:
"""

class InformationFunction(AbstractLogged):
    '''
    Abstract class for mapping hidden states to information sets
    '''
    def get_information_set(self, state: Hashable, actor: Hashable) -> Hashable:
        '''
        Returns the observed state (information set) for the state

        Args:
            state: current state
            actor: actor that is observing the state

        Returns:
            information_set: information set for the state
        '''
        return self._get_information_set(state=state, actor=actor)

    @abstractmethod
    def _get_information_set(self, state: Hashable, actor: Hashable) -> Hashable:
        '''
        Returns the observed state (information set) for the state

        Args:
            state: current state
            actor: actor that is observing the state

        Returns:
            information_set: information set for the state
        '''
        pass

"""
The player names are defined as follows:
{'guesser', 'clue_master'}
"""
\end{compressedlisting}

\begin{compressedlisting}[breakable, enhanced]{A: Action Function (Taboo)}
SYSTEM PROMPT:

"""
You are a programmer developing an accurate game engine. Your task is to implement parts of the game simulator in Python. The simulator models simultaneous actions as sequential ones with partial observation. When players $1, ..., k$ take simultaneous actions, they do so sequentially without seeing the actions of previous players. These actions are first recorded in a `HiddenState` object before being revealed. Do not repetitively generate Hidden state. If new state is generated, construct at return statement. `ObservedState` and `HiddenState` should be under `@dataclass(frozen=True)`.  Use tuple instead of list to make sure that vectors are frozen.
"""

HUMAN PROMPT:

"""
Taboo (2-player text version) is a two player cooperative dialogue game where 1 player is the clue-master and 1 player is the guesser. The clue master is given the clue-word and a list of taboo words. Each discussion round the clue master makes one statement to the guesser, but cannot use any of the taboo words in their statements. The guesser can then guess one word. This continues until either the guesser guesses the word, the guesser has guess five times already, or the clue-master has spoken one of the taboo words. If the clue-master uses any of the taboo words, the team score is 0. Otherwise, the score is five minus the number of words guesser has guessed.
"""

"""
A hidden state in the game is defined as follows:
"""
class HiddenState:
    from dataclasses import dataclass
    from typing import Optional, Tuple
    clue_word: str                               # The actual clue word (hidden from guesser)
    taboo_words: Tuple[str, ...]                 # Taboo words (hidden from guesser)
    taboo_word_used: bool                         # Whether a taboo word was used
    guesses: Tuple[str, ...]                      # Guesses made (hidden from clue-master)
    clue_master_statements: Tuple[str, ...]      # Statements made (hidden from guesser)
"""
Write an actor-action enumerator `CustomActorActionEnumerator` for this game that inherits from the `TextActorActionEnumerator` class. Include all docstings from the parent class:
"""
class TextActorActionEnumerator(ActorActionEnumerator):
    '''
    Abstract class for an actor action enumerator that can enumerate textual actions (such as dialogue and code)susing an LLM
    '''
    model: LLMModel

    def __init__(self, model: LLMModel, max_actions: int, player_order: Tuple[Hashable] = tuple()):
        super().__init__(player_order)
        self.model = model
        self.max_actions = max_actions

    def enumerate(self, state: Hashable)->tuple[Optional[Hashable], set]:
        '''
        Enumerates a (single) actor that may take actions at the state and the actions that the actor may take.
        If multiple actors may take actions at this state (simultaneous state), the first actor in the player order is returned.

        Args:
            state: current state

        Returns:
            actor: the acting actor. -1 for environment, None for terminal state
            actions: set of actions that the actor may take
        '''
        actor, actions = self._enumerate(state)
        assert len(actions) <= self.max_actions
        return actor, set(actions)

    @abstractmethod
    def _enumerate(self, state: Hashable)->tuple[Optional[Hashable], frozenset]:
        '''
        Enumerates a (single) actor that may take actions at the state and the actions that the actor may take.
        If multiple actors may take actions at this state (simultaneous state), the first actor in the player order is returned.

        Args:
            state: current state

        Returns:
            actor: the acting actor. -1 for environment, None for terminal state
            actions: set of actions that the actor may take

        For textual actions, the actions will be generated by prompting the LLM model with a system message and a user message, using the generate_k_responses method. An example of a system prompt is "You are the clue giver in the game of Codenames. The rules of Codenames are ...". An example of a user prompt is "State of game: ... Please give a clue as a single tuple (word, number), nothing else."
        '''
        pass

    def generate_k_responses(self, sys_prompt: SystemMessage, user_prompt: HumanMessage, k: int = -1)->list[str]:
        '''
        Generates k responses given the system prompt and user prompt

        Args:
            sys_prompt: system prompt
            user_prompt: user prompt
            k: number of responses to generate, -1 if set to self.max_actions

        Returns:
            responses: list of responses
        '''
        if k == -1:
            k = self.max_actions
        return [self.model.generate([sys_prompt, user_prompt]) for _ in range(k)]

"""
The player names are defined as follows:
{'guesser', 'clue_master'}
"""
\end{compressedlisting}

\begin{compressedlisting}{I: Information Realization Function (Taboo)}
SYSTEM PROMPT:

"""
You are a programmer developing an accurate game engine. Your task is to implement parts of the game simulator in Python. The simulator models simultaneous actions as sequential ones with partial observation. When players $1, ..., k$ take simultaneous actions, they do so sequentially without seeing the actions of previous players. These actions are first recorded in a `HiddenState` object before being revealed. Do not repetitively generate Hidden state. If new state is generated, construct at return statement. `ObservedState` and `HiddenState` should be under `@dataclass(frozen=True)`.  Use tuple instead of list to make sure that vectors are frozen.
"""
    
HUMAN PROMPT:

"""
Taboo (2-player text version) is a two player cooperative dialogue game where 1 player is the clue-master and 1 player is the guesser. The clue master is given the clue-word and a list of taboo words. Each discussion round the clue master makes one statement to the guesser, but cannot use any of the taboo words in their statements. The guesser can then guess one word. This continues until either the guesser guesses the word, the guesser has guess five times already, or the clue-master has spoken one of the taboo words. If the clue-master uses any of the taboo words, the team score is 0. Otherwise, the score is five minus the number of words guesser has guessed.
"""
"""
A hidden state in the game is defined as follows:
"""
class HiddenState:
    from dataclasses import dataclass
    from typing import Optional, Tuple
    clue_word: str                               # The actual clue word (hidden from guesser)
    taboo_words: Tuple[str, ...]                 # Taboo words (hidden from guesser)
    taboo_word_used: bool                         # Whether a taboo word was used
    guesses: Tuple[str, ...]                      # Guesses made (hidden from clue-master)
    clue_master_statements: Tuple[str, ...]      # Statements made (hidden from guesser)
"""
An observation in the game is defined as follows:
"""
@dataclass(frozen=True)
class ObservedState:
    clue_word: Optional[str]                     # The word the guesser needs to guess
    taboo_words: Optional[tuple[str, ...]] # List of taboo words the clue-master cannot use
    guesses: tuple[str, ...] = tuple()  # List of words guessed by the guesser
    clue_master_statements: tuple[str, ...] = tuple()  # Statements made by the clue-master
    taboo_word_used: bool = False      # Flag to indicate if a taboo word was used
    game_over: bool = False            # Flag to indicate if the game is over
    score: int = 5                     # Initial score, will decrease based on guesses
    actor: str = "clue_master"          # Indicates whose turn it is: "clue_master" or "guesser"

"""
Write an information prior `CustomInformationPrior` for this game that inherits from the `InformationPrior` class. Include all docstings from the parent class:
"""
class InformationPrior(AbstractLogged):
    '''
    Abstract class for mapping an information set to a hidden state.

    This is particularly useful when you do not have an empirical distribution over the hidden states for that information set
    '''
    def __init__(self, rng: np.random.Generator = np.random.default_rng()):
        self.rng = rng
        super().__init__()

    def get_prior_state(self, information_set: Hashable) -> Hashable:
        '''
        Returns the prior state for the information set. Can be stochastic
        '''
        return self._get_prior_state(information_set=information_set)

    @abstractmethod
    def _get_prior_state(self, information_set: Hashable) -> Hashable:
        '''
        Returns a hidden state corresponding to the given information set (observed state). Since an information set can often map to multiple hidden states, this function may return results stochastically. For states involving simultaneous actions, it defaults to returning the hidden state where no simultaneous actions have been taken yet.
        '''
        pass

"""
The player names are defined as follows:
{'guesser', 'clue_master'}
""" 
\end{compressedlisting}

\begin{compressedlisting}{S: Hidden States (Taboo)}
SYSTEM PROMPT:

"""
You are a programmer developing an accurate game engine. Your task is to implement parts of the game simulator in Python. The simulator models simultaneous actions as sequential ones with partial observation. When players $1, ..., k$ take simultaneous actions, they do so sequentially without seeing the actions of previous players. These actions are first recorded in a `HiddenState` object before being revealed. Do not repetitively generate Hidden state. If new state is generated, construct at return statement. `ObservedState` and `HiddenState` should be under `@dataclass(frozen=True)`.  Use tuple instead of list to make sure that vectors are frozen.
"""

HUMAN PROMPT: 

"""
Taboo (2-player text version) is a two player cooperative dialogue game where 1 player is the clue-master and 1 player is the guesser. The clue master is given the clue-word and a list of taboo words. Each discussion round the clue master makes one statement to the guesser, but cannot use any of the taboo words in their statements. The guesser can then guess one word. This continues until either the guesser guesses the word, the guesser has guess five times already, or the clue-master has spoken one of the taboo words. If the clue-master uses any of the taboo words, the team score is 0. Otherwise, the score is five minus the number of words guesser has guessed.
"""
"""
An observed state (information set) in the game is defined as follows:
"""
@dataclass(frozen=True)
class ObservedState:
    clue_word: Optional[str]                     # The word the guesser needs to guess
    taboo_words: Optional[tuple[str, ...]] # List of taboo words the clue-master cannot use
    guesses: tuple[str, ...] = tuple()  # List of words guessed by the guesser
    clue_master_statements: tuple[str, ...] = tuple()  # Statements made by the clue-master
    taboo_word_used: bool = False      # Flag to indicate if a taboo word was used
    game_over: bool = False            # Flag to indicate if the game is over
    score: int = 5                     # Initial score, will decrease based on guesses
    actor: str = "clue_master"          # Indicates whose turn it is: "clue_master" or "guesser"

"""
The player names are defined as follows:
{'guesser', 'clue_master'}
"""
\end{compressedlisting}

\begin{compressedlisting}{T: Transition-Reward Function (Taboo)}
SYSTEM PROMPT:

"""
You are a programmer developing an accurate game engine. Your task is to implement parts of the game simulator in Python. The simulator models simultaneous actions as sequential ones with partial observation. When players $1, ..., k$ take simultaneous actions, they do so sequentially without seeing the actions of previous players. These actions are first recorded in a `HiddenState` object before being revealed. Do not repetitively generate Hidden state. If new state is generated, construct at return statement. `ObservedState` and `HiddenState` should be under `@dataclass(frozen=True)`.  Use tuple instead of list to make sure that vectors are frozen.
"""

HUMAN PROMPT:

"""
Taboo (2-player text version) is a two player cooperative dialogue game where 1 player is the clue-master and 1 player is the guesser. The clue master is given the clue-word and a list of taboo words. Each discussion round the clue master makes one statement to the guesser, but cannot use any of the taboo words in their statements. The guesser can then guess one word. This continues until either the guesser guesses the word, the guesser has guess five times already, or the clue-master has spoken one of the taboo words. If the clue-master uses any of the taboo words, the team score is 0. Otherwise, the score is five minus the number of words guesser has guessed.
"""

"""
A hidden state in the game is defined as follows:
"""
class HiddenState:
    from dataclasses import dataclass
    from typing import Optional, Tuple
    clue_word: str                               # The actual clue word (hidden from guesser)
    taboo_words: Tuple[str, ...]                 # Taboo words (hidden from guesser)
    taboo_word_used: bool                         # Whether a taboo word was used
    guesses: Tuple[str, ...]                      # Guesses made (hidden from clue-master)
    clue_master_statements: Tuple[str, ...]      # Statements made (hidden from guesser)

"""
Write a forward transitor `CustomForwardTransitor` for this game that inherits from the `ForwardTransitor` class. Include all docstings from the parent class:
"""

class ForwardTransitor(ABC):
    '''
    Abstract class for a forward dynamics transition model
    '''
    @abstractmethod
    def _transition(self, state: Hashable, action: Hashable, actor: Hashable) -> Tuple[Hashable, dict[Hashable, float]]:
        '''

        Args:
            state: current state
            action: action taken by the actor
            actor: actor that is taking the action. -1 for environment

        Returns:
            next_state: next state
            rewards: reward of the transition for each player
        '''
        pass

    def transition(self, state: Hashable, action: Hashable, actor: Hashable)->Tuple[Hashable, dict[Hashable, float]]:
        '''
        Transits to the next state given the current state, actor, and action taken by the actor. Transitions are deterministic, with all randomness handled by the environment actor and the actions it takes. If multiple actors take actions at this state, record their actions down one transition step at a time.

        Args:
            state: current state
            action: action taken by the actor
            actor: actor that is taking the action. -1 for environment

        Returns:
            next_state: next state
            rewards: reward of the transition for each player

        Hint:
            Initialize the new_state at the point of returning, avoid creating the new_state copy prematurely.
        '''
        state, rewards = self._transition(state, action, actor)
        return state, rewards

"""
The player names are defined as follows:
{'guesser', 'clue_master'}
"""
\end{compressedlisting}
\newpage

\section{Language action examples}
\label{sec:language_action_details}
\begin{tcolorbox}[title=Example proposed possible dialogue actions]
    Clue word: barefoot\\

    Taboo word: shoes, socks, summer, beach\\

    Action 1: You might feel the ground directly under your feet when you don’t wear any footwear.\\

    Action 2: It's a way to enjoy nature by feeling the earth, grass, or sand without anything covering your feet.\\
\end{tcolorbox}

\newpage




\newpage















\end{document}